\documentclass[]{spie}  %>>> use for US letter paper
\usepackage{amsmath,amsfonts,amssymb}
\usepackage{graphicx}
\usepackage{booktabs}
\usepackage{xcolor}
\usepackage[colorlinks=true, allcolors=blue]{hyperref}

\title{Marker-Constrained Pose-Graph Correction for Cross-Platform Georeferencing in GNSS-Denied Environments}

\author[a]{Marco Giberna}
\author[a]{Jose Luis Sanchez Lopez}
\author[a,b]{Holger Voos}
\affil[a]{Automation and Robotics Research Group, Interdisciplinary Centre for Security, Reliability and Trust (SnT), University of Luxembourg, 1855 Luxembourg, Luxembourg}
\affil[b]{Faculty of Science, Technology and Medicine, University of Luxembourg, 4365 Esch-sur-Alzette, Luxembourg}

\authorinfo{Further author information: (Send correspondence to M.G.)\\M.G.: E-mail: marco.giberna@uni.lu}

\begin{document} 
\maketitle

\begin{abstract}
Autonomous operation in GNSS-denied environments requires heterogeneous mapping pipelines to maintain a consistent spatial reference. 
This paper presents a framework using camouflage-matched fiducial markers fabricated from Cholesteric Spherical Reflectors (CSRs) as pre-surveyed visual anchors. 
The anchors georeference both a lightweight LiDAR-odometry trajectory and a dense RTAB-Map reconstruction, allowing their outputs to be expressed in a common LUREF frame (geodetic coordinate reference system used in Luxembourg) without requiring GNSS measurements during operation. 
The method combines coarse similarity alignment with marker-constrained pose-graph optimization. 
We evaluate it using two handheld acquisition sessions with ground-level and elevated motion profiles emulating UGV and UAV operation. 
A single iMarker was relocated among six surveyed positions, with the first position revisited to quantify drift correction. 
Marker-anchor correction reduced revisit inconsistency by 97.9\% and 99.1\% for the UAV- and UGV-emulating sessions, respectively, and improved held-out anchor prediction compared with one-time alignment. 
Separately georeferenced dense reconstructions achieved a median cross-session nearest-neighbour distance of 58 cm without explicit cross-session registration. 
Marker processing operated in real time, while trajectory correction required less than 0.25 s per session. 
These results demonstrate a proof of concept for georeferencing lightweight odometry and dense reconstructions using visually unobtrusive, pre-surveyed anchors during GNSS-denied operation.
\end{abstract}


\section{Introduction}
\label{sec:introduction}

Autonomous multi-platform operations, combining Unmanned Aerial Vehicles (UAVs) and Unmanned Ground Vehicles (UGVs), are increasingly central to defence, border protection, and reconnaissance missions. 
These operations frequently take place in GNSS-denied environments, whether due to natural signal degradation, evolving dense urban or forested terrain, or deliberate adversarial interference\cite{giberna2026digital}. 
In such settings, maintaining consistent spatial awareness across heterogeneous platforms operating at different altitudes, scales, and sensing modalities becomes a critical bottleneck for collaborative mission execution.

Robotic mapping pipelines commonly produce outputs with different purposes and computational characteristics~\cite{campos2021orbslam3, labbe2019rtabmap, giberna2025dynemo}. 
Lightweight odometry methods provide real-time pose estimates suitable for onboard operation but may accumulate substantial drift when they lack loop closure or global optimization. 
Dense mapping systems integrate RGB-D or LiDAR measurements into detailed environmental reconstructions and may apply loop-closure constraints to improve global consistency. 
Although these outputs may describe the same environment, they are not automatically expressed in a common geodetic frame, particularly when GNSS is unavailable during operation. 
This limits map comparison, information exchange, and localization across sessions or platforms.

% A central challenge in this context is the representational gap between dense mapping and sparse Simultaneous Localization and Mapping (SLAM). 
% Dense 2D or 3D maps provide the rich environmental detail needed for situational awareness, obstacle avoidance, and mission planning, while SLAM-based localization typically relies on sparse keypoint representations optimized for real-time onboard performance~\cite{campos2021orbslam3, labbe2019rtabmap, giberna2025dynemo}. These two representations are difficult to fuse directly across platforms: a UAV's sparse visual-inertial trajectory and a UGV's dense reconstructed map do not share a common, reliably matchable structure, especially without GNSS to anchor both to a global frame.

Fiducial markers have long been used to inject reliable, artificial structure into visual localization pipelines, offering known geometry and unambiguous identification that natural scene features cannot guarantee~\cite{garridojurado2014aruco, olson2011apriltag}. 
However, conventional markers are visually conspicuous: a significant limitation for defence and security applications where undisguised markers would compromise concealment, or where environmental and operational constraints prohibit visible artificial infrastructure.

This paper introduces a framework built around imperceptible fiducial markers (iMarkers), fabricated from Cholesteric Spherical Reflectors (CSRs)~\cite{agha2022unclonable}. 
iMarkers selectively reflect narrow, non-visible or near-visible wavelength bands (NIR, NUV, or visible), remain invisible to the naked eye, and can be fully camouflaged into arbitrary surfaces while remaining reliably detectable to appropriately equipped sensors~\cite{agha2022unclonable, tourani2024imarkers}. 
Their spherical geometry ensures omnidirectional reflection independent of viewing angle, and their circularly polarized reflection response allows clean separation from background clutter, properties that make them robust under the variable lighting and atmospheric conditions typical of field deployment, and specifically well-suited to concealed use in defence and security contexts.

We propose using strategically placed iMarkers as secure, reliable visual anchor points that provide common geodetic anchors for a lightweight LiDAR-odometry trajectory and a dense RTAB-Map reconstruction. 
By relating marker observations to coordinates surveyed in advance, the framework places the resulting trajectory and dense reconstruction in a common geodetic frame without requiring GNSS availability during data acquisition or online localization.

The contributions of this paper are as follows:
\begin{itemize}
    \item We present a marker-anchor framework that places a lightweight LiDAR-odometry trajectory and a dense RTAB-Map reconstruction in a common pre-surveyed reference frame during GNSS-denied operation.
    \item We describe the integration of camouflage-matched CSR iMarkers with RGB-based detection, PnP pose estimation, similarity alignment, and marker-constrained pose-graph optimization.
    \item We evaluate the method on UAV- and UGV-emulating acquisition sessions, quantifying trajectory correction, held-out georeferencing, marker-pose repeatability, anchor-placement effects, cross-session reconstruction agreement, and computational cost.
\end{itemize}

% The remainder of this paper is organized as follows. Section~\ref{sec:related_work} reviews related work in fiducial marker systems and cross-platform SLAM fusion. Section~\ref{sec:imarker_design} describes the design and fabrication of iMarkers. Section~\ref{sec:architecture} presents the system architecture bridging dense and sparse representations. Sections ~\ref{sec:results} and ~\ref{sec:discussion} present results and discussion.

\section{Related Work}
\label{sec:related_work}

\subsection{Fiducial Marker Systems}
Fiducial markers such as ArUco~\cite{garridojurado2014aruco} and AprilTag~\cite{olson2011apriltag} are widely adopted in robotics for providing unambiguous, geometrically precise reference points, supporting applications from camera calibration to relative pose estimation. 
These systems are effective but rely on high-contrast, visually distinct printed patterns, which limits their applicability where visual concealment is required or where marker presence would interfere with the operational or aesthetic requirements of the deployment environment.

\subsection{Cholesteric Spherical Reflectors and iMarkers}
Cholesteric Spherical Reflectors (CSRs) offer optical properties fundamentally different from conventional retroreflective or printed markers: omnidirectional reflection due to their spherical geometry, wavelength-selective reflection enabling operation in NIR/NUV or narrow visible bands, and circularly polarized reflection response that supports robust separation from ambient background signals~\cite{agha2022unclonable}. 
These properties collectively enable markers that are imperceptible to human observers while remaining consistently detectable to appropriately filtered sensor, a combination not achievable with conventional fiducial systems. 
Building on this foundational CSR technology, subsequent work has integrated these markers, termed iMarkers, into robotic perception pipelines, including semantic scene graph construction~\cite{tourani2024imarkers} and broader robotic mapping applications~\cite{tourani2025imarkers}, demonstrating their practical viability as an alternative to visually obtrusive fiducial systems.

\subsection{Lightweight Odometry and Dense Mapping}
Lightweight state-estimation pipelines prioritize real-time pose estimation using compact state representations and sparse sensor observations. Feature-based visual SLAM, visual-inertial estimation, and LiDAR odometry are representative examples, although their outputs and global-consistency mechanisms differ substantially\cite{geneva2020openvins, xu2022fast}. 
When loop closure or global optimization is absent, the resulting trajectories may accumulate drift over time. 
Dense mapping approaches instead integrate imagery, depth, or LiDAR measurements into detailed environmental reconstructions, often at greater computational cost. 
RTAB-Map~\cite{labbe2019rtabmap} supports multimodal dense reconstruction together with loop-closure-based trajectory optimization. 
In multi-session or heterogeneous-platform operation, the relevant challenge is therefore not necessarily direct fusion between a sparse landmark map and a dense map, but establishing a common reference for different trajectory and reconstruction outputs.

% Dense mapping approaches provide detailed environmental representations well-suited to navigation and situational awareness, but are computationally costly to maintain in real time and difficult to align directly across independently operating platforms. 
% Sparse SLAM systems, including feature-based methods such as ORB-SLAM3~\cite{campos2021orbslam3}, visual-inertial approaches such as VINS-Fusion~\cite{qin2019vinsfusion}, and filter-based frameworks such as OpenVINS~\cite{geneva2020openvins}, prioritize efficient, real-time pose estimation using sparse keypoints. 
% RTAB-Map~\cite{labbe2019rtabmap} notably supports both dense reconstruction and loop-closure-based SLAM within a single framework, fusing RGB-D, LiDAR, and inertial data. 
% Despite these advances, sparse SLAM maps remain often insufficient as a basis for fusing with another platform's dense representation, particularly in GNSS-denied settings where no common global reference is otherwise available.

\subsection{Cross-Platform Map Alignment and GNSS-Denied Georeferencing}
Multi-robot SLAM systems commonly merge locally constructed maps by detecting inter-robot place correspondences and introducing cross-robot loop-closure constraints into a centralized or distributed optimization. 
CCM-SLAM\cite{schmuck2019ccm} performs centralized place recognition, map fusion, and global optimization for multiple monocular agents, whereas Kimera-Multi\cite{tian2022kimera} uses distributed place recognition and robust pose-graph optimization to fuse local trajectories and dense metric-semantic maps. 
These approaches depend on reliable cross-map data association from overlapping observations.
This requirement becomes more difficult for heterogeneous aerial–ground sensing, where measurements can differ substantially in viewpoint, point density, and sensing characteristics. 
CrossLoc3D\cite{guan2023crossloc3d} formulates aerial–ground LiDAR place recognition as a cross-source representation-gap problem involving different perspectives, point densities, and noise characteristics, while Gao et al.\cite{gao2023visual} address cross-view and cross-modal place recognition for aerial–ground collaborative perception.
Engineered fiducial landmarks provide an alternative source of explicit and repeatable correspondence\cite{pfrommer2019tagslam}. 
Building on this principle, our work uses pre-surveyed iMarkers to relate the ICP-odometry and RTAB-Map outputs to the same LUREF frame without requiring GNSS measurements during operation or direct natural-feature matching between the two representations.

% In the absence of GNSS, establishing a reliable common reference frame for georeferencing becomes particularly difficult, motivating the use of engineered, unambiguous anchor points rather than relying solely on naturally occurring scene structure. 
% Our work uses iMarkers as engineered visual anchors that can be observed by both processing pipelines, allowing their outputs to be related to a common pre-surveyed frame without GNSS availability during operation or extensive shared scene overlap.

\section{iMarker Design and Fabrication}
\label{sec:imarker_design}
\subsection{CSR-Based Marker Principle}
iMarkers are fabricated from Cholesteric Spherical Reflectors (CSRs), micro-scale spherical structures that exploit the chiral Bragg diffraction of cholesteric liquid crystals to achieve omnidirectional, wavelength-selective reflection~\cite{agha2022unclonable}. 
Unlike planar retroreflective materials, the spherical geometry of CSRs guarantees a consistent reflection response regardless of the viewing angle between sensor and marker, while their circularly polarized reflection allows detection systems to discriminate marker signal from unpolarized ambient background clutter. 
This combination of properties has previously been leveraged to produce fiducial markers that are difficult to perceive under normal viewing conditions while remaining reliably detectable to matched sensing and processing pipelines~\cite{agha2022unclonable, tourani2025imarkers}.

\subsection{Wavelength Selection: Visible-Light, Camouflage-Matched Design}
CSR-based markers can be formulated to reflect selectively in the near-infrared (NIR), near-ultraviolet (NUV), or visible spectrum. Prior work has predominantly emphasized NIR or NUV formulations, where imperceptibility is achieved through wavelength selection outside the range of human visual sensitivity~\cite{agha2022unclonable, tourani2025imarkers}. 
In this work, we instead adopt a visible-light, green-reflective CSR formulation, with imperceptibility achieved through a complementary mechanism: color-matching the marker to a camouflaged substrate on which it is deployed, rather than through wavelength-based invisibility.

\begin{figure} [ht]
\begin{center}
\begin{tabular}{c} 
\includegraphics[width=\textwidth]{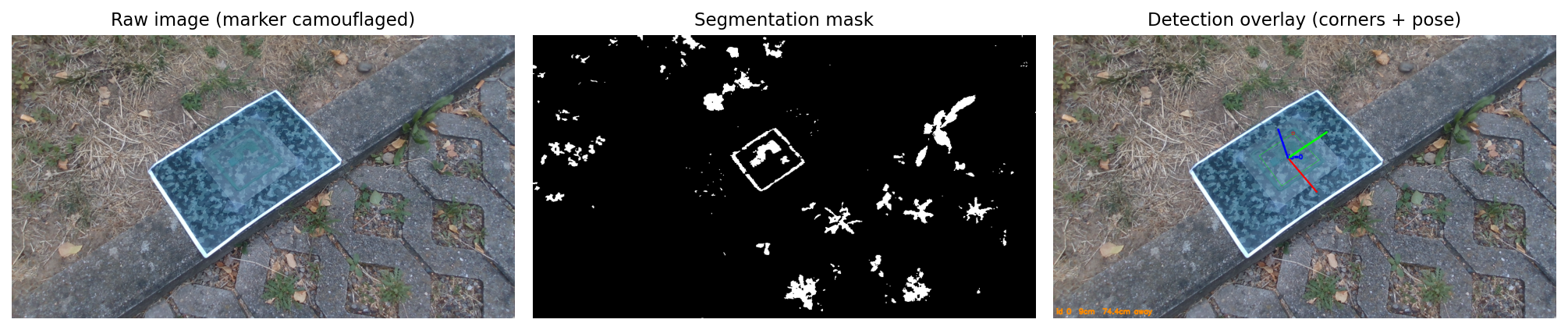}
\end{tabular}
\end{center}
\caption[iMarker] 
{ \label{fig:iMarker} 
Detection pipeline output on a single frame. Left: raw image, marker camouflaged against its background. Centre: segmentation mask isolating the CSR-reflective region. Right: detected marker with recovered pose overlaid.}
\end{figure}

This design choice has practical motivation relevant to the defence deployment context targeted by this work. 
First, it allows detection with conventional RGB sensing on the same platform used for dense mapping, simplifying the sensor payload. 
Second, it aligns naturally with existing camouflage doctrine and materials already used in defence and security contexts, where markers can be integrated into camouflage netting, painted equipment, or vegetation-toned surfaces without introducing a new spectral signature that specialized detection equipment (e.g., NIR-sensitive imaging) could exploit to reveal marker presence.
Figure \ref{fig:iMarker} shows an example of iMarker employed during data collection.

\subsection{Marker Geometry and Mask Design}
Each iMarker encodes a fixed geometric pattern, following an AprilTag-style encoding scheme~\cite{olson2011apriltag}, to support recognition and pose estimation once the CSR region has been isolated from the surrounding scene. 
The physical dimensions of the marker were measured and documented precisely, enabling model-based pose estimation via a Perspective-n-Point (PnP) formulation using the marker's known real-world geometry together with calibrated camera intrinsics.

Detection follows a two-stage process consistent with the iMarker detection framework of Tourani et al.\cite{tourani2025imarkers} : a mask generation stage isolates the CSR-reflective region of the image based on its distinctive spectral/polarization response, followed by ArUco-style pattern recognition applied to the mask to confirm marker identity and recover its corner geometry for pose estimation.
Figure \ref{fig:iMarker} depicts a this process on a iMarker image frame.

\section{System Architecture}
\label{sec:architecture}

\subsection{Overview}
\begin{figure} [ht]
\begin{center}
\begin{tabular}{c} 
\includegraphics[width=0.8\textwidth]{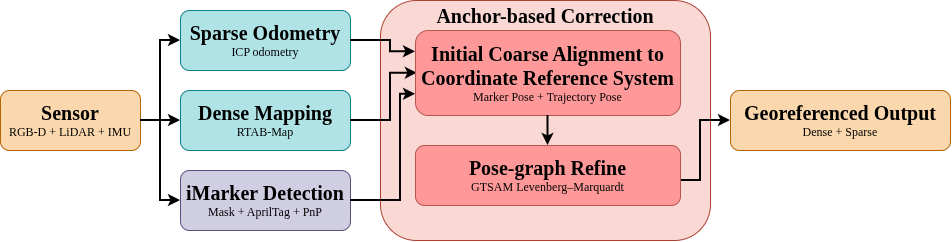}
\end{tabular}
\end{center}
\caption[system architecture] 
{ \label{fig:architecture} 
System architecture. The sensor device feeds lightweight ICP-based LiDAR odometry and iMarker detection. ICP odometry initializes dense mapping (RTAB-Map), which optimizes its own trajectory via loop closure. Both trajectories and marker detections feed the anchor-based correction module, which coarsely aligns and then pose-graph-refines each trajectory into the LUREF frame.}
\end{figure}

The proposed framework consists of four functional components operating on data from a multi-sensor device\cite{bastos2026smapper}: (i) a \textit{sparse ICP-based LiDAR odometry pipeline}, (ii) a \textit{dense mapping pipeline} initialized using the ICP trajectory, (iii) an \textit{iMarker detection and correspondence module}, and (iv) an \textit{anchor-based correction module}, applied independently to both the ICP and RTAB-Map trajectories. 
Figure~\ref{fig:architecture} illustrates the overall data flow. 
This composition, dense mapping seeded by sparse odometry, both independently corrected via the same marker anchors, differs from the originally-intended fully-decoupled dense/sparse architecture, and the reasons for this composition (Section~\ref{sec:results}) are themselves a contribution of this work.

\subsection{Sensor Input}
The framework operates on synchronized RGB-D, LiDAR, and inertial data captured from a single handheld sensor device\cite{bastos2026smapper}. 
UAV and UGV platform behavior is emulated by varying the trajectory profile with which this rig is moved through the environment.

\subsection{Lightweight LiDAR Odometry Pipeline}
The odometry-side output is produced using ICP odometry, a loosely coupled frame-to-model LiDAR scan-matching method available in the RTAB-Map ecosystem~\cite{labbe2019rtabmap}.
This configuration was selected empirically: several visual-inertial and LiDAR-inertial alternatives were evaluated and found unsuitable for the motion profile characteristic of this study's recordings, whereas loosely-coupled LiDAR-only odometry proved substantially more robust. 
This pipeline has no loop-closure or global consistency mechanism: its trajectory estimate accumulates drift monotonically over a session with no means of self-correction.
Its output is a drift-prone metric pose trajectory rather than a sparse landmark map, representing the type of lightweight state estimate that could be maintained onboard a resource-constrained platform.

\subsection{Dense Mapping Pipeline}
Dense mapping is performed using RTAB-Map~\cite{labbe2019rtabmap}, configured to fuse RGB-D imagery and LiDAR scans for dense 3D reconstruction and loop-closure detection. 
To obtain full-session dense coverage, RTAB-Map's mapping node is decoupled from its own odometry front-end and supplied with \textit{ICP odometry}'s trajectory as an externally-provided pose source, using RTAB-Map exclusively for dense reconstruction and its independent RGB-D+LiDAR loop-closure detection. 
Although RTAB-Map is initialized using the ICP-odometry poses, it subsequently optimizes a separate trajectory through its RGB-D and LiDAR loop-closure constraints. The resulting trajectory is therefore related to, but not identical to, the ICP trajectory used as its initial pose source.

\subsection{iMarker Detection Module}
The RGB stream is processed by the iMarker detection module, built on the iMarker detection framework~\cite{tourani2025imarkers}. 
This module isolates CSR-reflective regions via HSV-based mask generation, applies AprilTag-style pattern recognition~\cite{olson2011apriltag} to the masked region, and recovers marker corner geometry. 
Using the marker's known physical dimensions together with calibrated camera intrinsics, a Perspective-n-Point (PnP) solve yields a full 6-DoF marker pose in the camera frame for each positive detection.

% \textcolor{red}{TODO: consider removing if removed before}
% Because only a single physical marker was fabricated and relocated sequentially between surveyed locations (Section~\ref{sec:imarker_design_challenges}), detections cannot be associated with a known location via decoded marker identity. 
% Correspondence is instead established through temporal gap-clustering of detections within each recording session: detections are grouped into clusters separated by transit gaps, and clusters are labeled in the session's known visitation order. 
% Across both evaluated sessions, this reliably recovers all seven expected clusters, six surveyed locations plus a revisit of the first, once a resource-management defect in the detection node (Section~\ref{sec:results}) was identified and corrected.

\subsection{Anchor-Based Trajectory Correction}
\label{sec:architecture_correction}
Each temporally-labeled marker detection cluster corresponds to a known position in the LUREF geodetic reference frame. 
Rather than using these correspondences only for a single post-hoc alignment, this framework uses them as \textit{correction constraints} within a pose-graph optimization over the ICP-odometry trajectory, implemented using GTSAM~\cite{factor_graphs_for_robot_perception}.
The same correction procedure is applied separately to the ICP-odometry trajectory and to the RTAB-Map optimized trajectory associated with the dense reconstruction.

The correction proceeds in two composed steps. 
First, a coarse similarity transform (rotation, translation, and scale) is fitted between the sparse trajectory's local frame and the LUREF frame using the Umeyama method over the available marker anchors, bringing the trajectory into approximate global alignment. 
Second, a pose graph is constructed from the trajectory's own consecutive-pose odometry edges, augmented with ground-control-point constraint edges at each marker anchor tying the corresponding trajectory node to its known LUREF position; the graph is optimized via Levenberg-Marquardt. 
This two-step composition is necessary: the pose-graph refinement alone, without prior coarse alignment, cannot correct the large initial frame/scale mismatch between the sparse trajectory's local frame and LUREF.

The odometry-edge noise model, required by the optimizer to weigh trajectory smoothness against anchor constraints, is calibrated empirically from directly observed trajectory drift (Section~\ref{sec:results}) rather than assumed, since the underlying LiDAR odometry does not publish a covariance estimate.

This anchor-based correction serves two complementary roles: (i) it substantially reduces accumulated trajectory drift, most directly demonstrated where a location is revisited later in a session, allowing the two independent visits to the same physical point to be reconciled; and (ii) it improves the trajectory's accuracy at held-out (non-anchor) points to a degree evaluated via leave-one-out cross-validation, though this improvement is shown to depend on the local temporal density of surrounding anchors rather than being uniform across the trajectory (Section~\ref{sec:results}).

\subsection{Output}
The framework outputs a corrected, LUREF-georeferenced ICP-odometry trajectory; a separately corrected RTAB-Map trajectory and its associated dense reconstruction; and the marker observations used for anchor correction and evaluation.

\section{Results}
\label{sec:results}
The framework was evaluated on two full recording sessions, a ground-level ("UGV") trajectory of 362.4\,s and an elevated ("UAV") trajectory of 319.1\,s, each visiting six surveyed LUREF locations plus a revisit of the first (coordinates are reported in Table \ref{tab:repeatability}), using a single physical iMarker relocated sequentially between locations. Marker detections were grouped into seven temporal clusters per session (A--F, plus a revisit A$_{\text{revisit}}$) via gap-based segmentation on recording-session metadata, since location correspondence cannot rely on decoded marker identity.

\subsection{Within-Cluster Marker-Pose Repeatability}
Absolute per-detection error against the surveyed LUREF coordinates cannot isolate the accuracy of the marker detector or PnP pose estimator, because each marker estimate is composed through the estimated sensor trajectory. 
It therefore also reflects trajectory interpolation, temporal synchronization, sensor extrinsic calibration, marker-placement uncertainty, and survey uncertainty. 
We consequently evaluate the empirical repeatability of the composed marker-position estimates within each detection cluster, rather than interpreting their absolute LUREF error as marker-detection accuracy.

For each cluster, we compute the centroid of its composed marker-position estimates and report the RMS distance of individual detections from that centroid (Table~\ref{tab:repeatability}).
Because the cluster's own centroid is removed, this measure is insensitive to cluster-level systematic offsets arising from global trajectory alignment, survey error, or marker placement; it retains only short-term scatter from trajectory interpolation, synchronization, calibration, detection, and PnP estimation. 
It should therefore be read as end-to-end local repeatability rather than absolute positional accuracy.

\begin{table}[h]
\centering
\caption{Per-cluster marker-pose repeatability: RMS distance of individual detections from
their cluster's own centroid, cm, alongside each cluster's surveyed LUREF location.}
\label{tab:repeatability}
\begin{tabular}{lcccccccc}
\toprule
& \multicolumn{3}{c}{LUREF coordinates} & \multicolumn{2}{c}{UAV} & \multicolumn{2}{c}{UGV} \\
Cluster & N & E & h & $n$ & RMS & $n$ & RMS \\
\midrule
A & 76839.129 & 79345.498 & 340.570 & 152 & 14.71 & 226 & 13.76 \\
B & 76829.898 & 79357.502 & 339.409 & 289 & 22.54 & 84 & 37.55 \\
C & 76821.262 & 79363.262 & 338.276 & 167 & 10.94 & 260 & 20.43 \\
D & 76789.028 & 79320.063 & 338.382 & 209 & 6.03 & 92 & 8.11 \\
E & 76804.525 & 79309.431 & 340.234 & 197 & 18.21 & 200 & 15.44 \\
F & 76815.709 & 79296.809 & 341.129 & 185 & 14.43 & 110 & 22.70 \\
A$_{\text{revisit}}$ & 76839.129 & 79345.498 & 340.570 & 175 & 7.94 & 181 & 12.22 \\
\midrule
\textbf{Overall} & & & & \textbf{1374} & \textbf{15.31} & \textbf{1153} & \textbf{18.79} \\
\bottomrule
\end{tabular}
\end{table}

Pooled within-cluster repeatability is 15.31\,cm (UAV) and 18.79\,cm (UGV). 

\subsection{Georeferencing Accuracy and the Contribution of Anchor Correction}
Table~\ref{tab:correction_summary} summarizes the framework's central result: the effect of using marker anchors as active correction constraints (Section~\ref{sec:architecture_correction}), rather than only as a one-time reference, on both self-consistency at a revisited location and held-out generalization accuracy.

\begin{table}[!ht]
\centering
\caption{Effect of anchor-based correction, both trajectories. Self-consistency: distance
between two independent visits to the same LUREF location (A--A$_{revisit}$), cm. Generalization:
leave-one-out mean/RMS error over held-out anchors, cm.}
\label{tab:correction_summary}
\begin{tabular}{@{}ll@{}cc|cc}
\toprule
& & \multicolumn{2}{c|}{Self-consistency (revisit)} & \multicolumn{2}{c}{Generalization (LOO)} \\
Session & Trajectory & Initial Alignment & Pose-Graph Refined & Initial Alignment & Pose-Graph Refined \\
\midrule
UAV & ICP odometry & 942.0 & \textbf{20.0} & 215.9 / 231.9 & \textbf{169.6 / 181.9} \\
UAV & Dense mapping & 763.1 & \textbf{8.7} & 187.7 / 200.6 & \textbf{131.9 / 139.7} \\
\midrule
UGV & ICP odometry & 10{,}102.0 & \textbf{91.6} & 1297.6 / 1485.4 & \textbf{1193.1 / 1362.4} \\
UGV & Dense mapping & 12.9 & \textbf{10.9} & 533.2 / 662.1 & \textbf{419.9 / 516.8} \\
\bottomrule
\end{tabular}
\end{table}

Anchor-based correction reduces self-consistency error substantially for both trajectories: by 97.9\% (UAV ICP odometry), 98.9\% (UAV dense mapping), and 99.1\% (UGV ICP odometry).
The UGV dense trajectory shows a smaller relative reduction (15.5\%) because its raw, uncorrected self-consistency (12.9\,cm) was already close to the corrected result for the other three cases, leaving little room for further improvement; in absolute terms it remains the most self-consistent of the four uncorrected trajectories by a wide margin, consistent with the dense mapping pipeline's own independent loop-closure mechanism.
Held-out generalization accuracy improves by 8--30\% across the four session/trajectory combinations (8--9\% for ICP odometry, 21--30\% for dense mapping) over a coarse one-time alignment alone. 
The smaller relative improvement for the ICP-odometry trajectories, particularly UGV, is attributable to the correction procedure's odometry-edge noise model, which is calibrated once from observed drift and, in this case, under-corrects for a higher true drift rate on the longer UGV trajectory (Section~\ref{sec:discussion}); this parameter was deliberately not re-tuned per result to avoid overfitting the reported figures.

\begin{figure} [ht]
\begin{center}
\begin{tabular}{c} 
\includegraphics[width=0.9\textwidth]{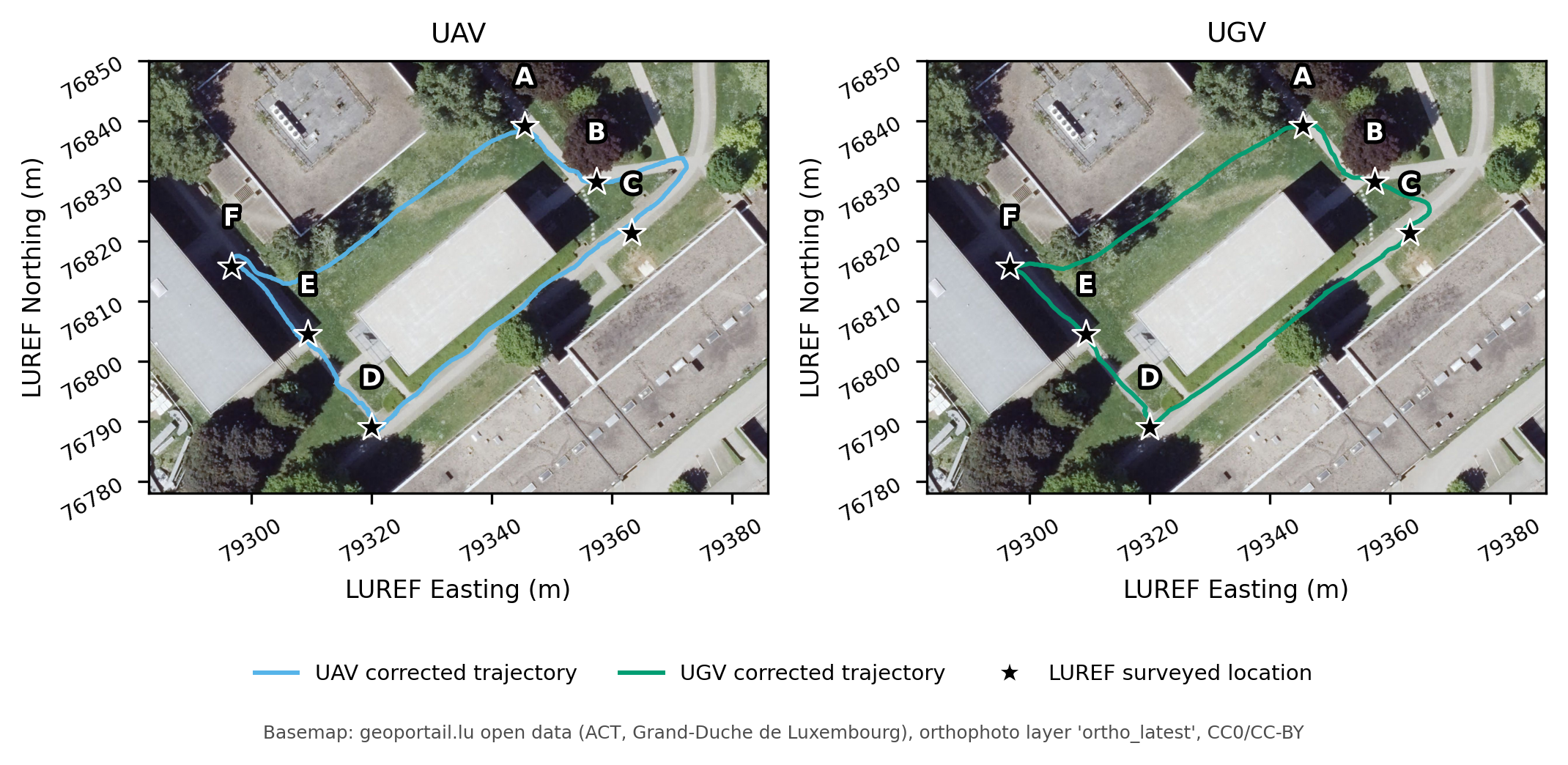}
\end{tabular}
\end{center}
\caption[trajectory map] 
{ \label{fig:trajectory_map} 
Pose-graph-corrected ICP-odometry trajectories for the UAV (left) and UGV (right) sessions, overlaid on a georeferenced orthophoto in LUREF coordinates. Stars mark the six surveyed anchor locations (A--F); labels indicate the corresponding temporal detection cluster. Both trajectories trace the same deployment loop and pass close to every surveyed location.}
\end{figure}

\begin{figure} [ht!]
\centering
\begin{tabular}{c} 
\multicolumn{1}{c}{\textbf{ICP Odometry}} \\[0.5em]
\includegraphics[width=0.8\textwidth]{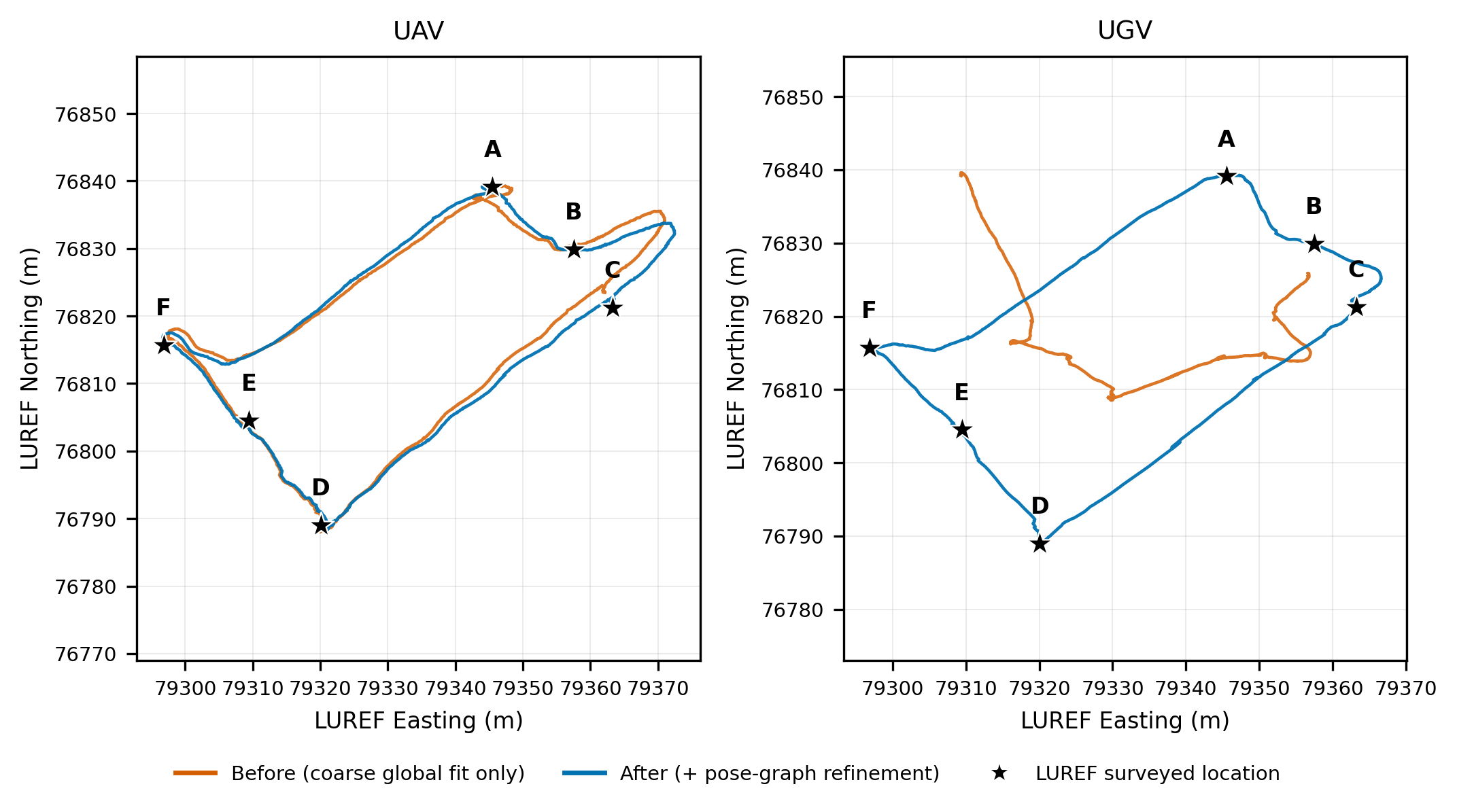} \\[1em]
\multicolumn{1}{c}{\textbf{Dense Mapping}} \\[0.5em]
\includegraphics[width=0.8\textwidth]{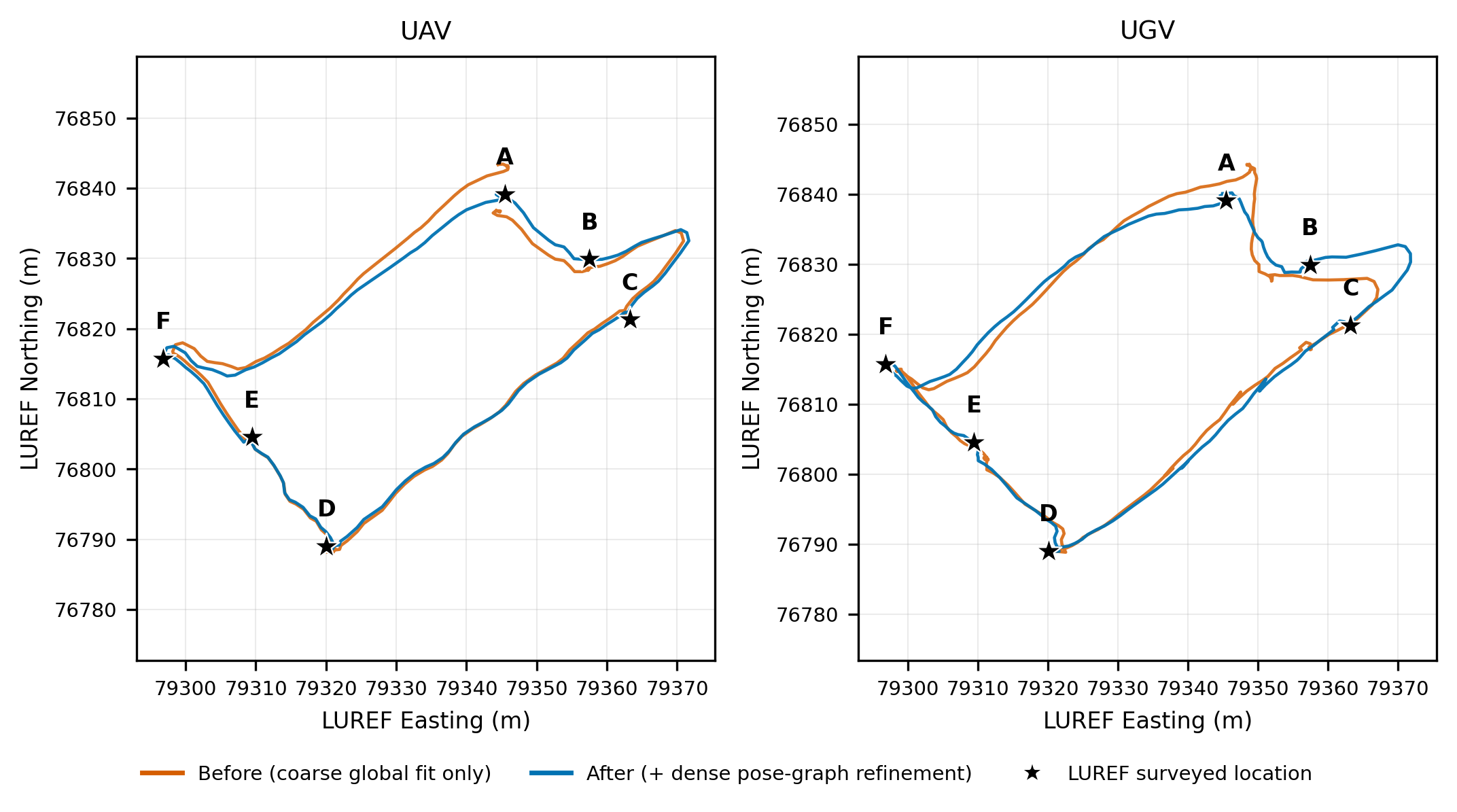}
\end{tabular}
\caption[correction before after] 
{ \label{fig:correction_before_after} 
ICP-odometry (top) and dense-mapping (bottom) trajectories for the UAV and UGV sessions, before (coarse alignment only) and after anchor-based pose-graph correction. Stars mark the six surveyed anchor locations.}
\end{figure}

Figure~\ref{fig:trajectory_map} shows the corrected trajectories overlaid on a georeferenced basemap alongside the surveyed marker locations, and Figure~\ref{fig:correction_before_after} shows the trajectory before and after anchor-based correction directly, illustrating the improvement summarized in Table~\ref{tab:correction_summary}.

\subsection{Accuracy as a Function of Anchor Count}
Table~\ref{tab:anchor_density} reports how georeferencing accuracy varies with the number of marker anchors available for the alignment fit, for both trajectories, averaged over the two sessions. 
For each anchor count $k < 6$, all $\binom{6}{k}$ possible subsets of the six surveyed anchors were used to fit a similarity transform, and the resulting held-out error on the remaining anchors was averaged across all subsets; $k=5$ is equivalent to, and reproduces, the leave-one-out result reported in Section~\ref{sec:results}. 

\begin{table}[h]
\centering
\caption{In-sample vs.\ held-out georeferencing accuracy as a function of anchor count $k$,
mean/RMS cm, averaged over both sessions. Held-out values are averaged over all
$\binom{6}{k}$ subsets.}
\label{tab:anchor_density}
\begin{tabular}{lcccc}
\toprule
Trajectory & $k$ & In-sample & Held-out \\
\midrule
ICP odometry & 6 & 475.1 / 526.8 & -- \\
ICP odometry & 5 & 457.6 / 494.1 & 752.8 / 854.2 \\
ICP odometry & 4 & 412.0 / 443.0 & 822.0 / 909.9 \\
ICP odometry & 3 & 281.9 / 340.2 & 1287.7 / 1945.7 \\
\midrule
Dense mapping & 6 & 226.6 / 269.3 & -- \\
Dense mapping & 5 & 214.8 / 253.6 & 356.5 / 426.3 \\
Dense mapping & 4 & 190.2 / 228.6 & 383.2 / 453.6 \\
Dense mapping & 3 & 132.5 / 177.4 & 749.9 / 1563.4 \\
\bottomrule
\end{tabular}
\end{table}
% \begin{table}[h]
% \centering
% \caption{In-sample vs.\ held-out georeferencing accuracy as a function of anchor count $k$,
% mean/RMS cm, on the dense mapping trajectory. Held-out values are averaged over all subsets.}
% \label{tab:anchor_density}
% \begin{tabular}{lcccc}
% \toprule
% Session & $k$ & Subsets & In-sample & Held-out \\
% \midrule
% UAV & 6 & 1 & 105.4 / 113.1 & -- \\
% UAV & 5 & 6 & 98.8 / 105.2 & 183.2 / 195.3 \\
% UAV & 4 & 15 & 85.2 / 92.3 & 201.6 / 217.3 \\
% UAV & 3 & 20 & 55.4 / 65.5 & 413.4 / 684.7 \\
% \midrule
% UGV & 6 & 1 & 347.8 / 425.5 & -- \\
% UGV & 5 & 6 & 330.8 / 401.9 & 529.9 / 657.3 \\
% UGV & 4 & 15 & 295.2 / 364.9 & 564.8 / 689.8 \\
% UGV & 3 & 20 & 209.6 / 289.2 & 1086.5 / 2442.1 \\
% \bottomrule
% \end{tabular}
% \end{table}

Held-out accuracy degrades monotonically as anchor count is reduced, for both trajectories, confirming that georeferencing quality depends directly on the number of independent locations at which a marker anchor is available. 
The dense-mapping trajectory is consistently more accurate than ICP odometry at every anchor count, consistent with the correction results in Table~\ref{tab:correction_summary}, but both trajectories show the same qualitative pattern: held-out RMS increases by roughly $2.3\times$ (ICP odometry) and $2.5\times$ (dense mapping) from $k=5$ to $k=3$. 
At the minimum viable anchor count ($k=3$), held-out error reaches the multi-metre range for both trajectories, and should be understood as providing a coarse geolocation only, not the accuracy achievable with denser coverage.

\subsection{Effect of Anchor Geometric Distribution}
Beyond anchor count, the spatial arrangement of anchors relative to one another was examined. 
Markers B and E are each positioned close to the straight line connecting their respective neighboring anchor pairs (A--C for B, D--F for E). 
Measured directly from surveyed LUREF coordinates, B lies 201.5\,cm (7.96\% of the A--C segment length) from that line, and E lies 220.4\,cm (6.21\% of the D--F segment length) from the D--F line, both near the segment midpoint. 
For comparison, every other anchor's distance from its own alternating-neighbor line (e.g.\ A relative to the F--B line) is 5--9$\times$ larger (1014--1803\,cm), confirming B and E are genuinely, quantifiably closer to collinear with their neighbors than any other anchor pair in this deployment. See Figures \ref{fig:trajectory_map} and \ref{fig:correction_before_after} for reference.

\begin{table}[h]
\centering
\caption{Mean held-out error for the near-collinear anchors \{B, E\} vs.\ the remaining
anchors \{A, C, D, F\}, pose-graph leave-one-out evaluation, before and after refinement.
Errors in cm.}
\label{tab:collinearity}
\begin{tabular}{llcccc}
\toprule
Session & Trajectory & Stage & Mean \{B, E\} & Mean \{A, C, D, F\} & Ratio \\
\midrule
UAV & ICP odometry & Coarse & 166.1 & 549.3 & 0.30 \\
UAV & ICP odometry & Refined & 98.6 & 205.0 & 0.48 \\
UGV & ICP odometry & Coarse & 1870.8 & 2485.5 & 0.75 \\
UGV & ICP odometry & Refined & 279.1 & 1650.1 & 0.17 \\
UAV & Dense mapping & Coarse & 170.5 & 218.7 & 0.78 \\
UAV & Dense mapping & Refined & 101.4 & 147.2 & 0.69 \\
UGV & Dense mapping & Coarse & 535.4 & 523.8 & 1.02 \\
UGV & Dense mapping & Refined & 276.2 & 491.8 & 0.56 \\
\bottomrule
\end{tabular}
\end{table}

The expected consequence of this near-collinear placement would be a geometrically weaker contribution to the alignment fit.
Testing this directly across twelve independent leave-one-out evaluations (both trajectories, both sessions, both fit types), B and E instead show \emph{lower} held-out error than the remaining anchors in eleven of twelve cases, typically by a factor of 2--5$\times$.
This is the opposite of the naive expectation, and is attributable to a different mechanism than geometric fit weakness: a near-collinear anchor bracketed by two well-separated, retained neighbors is effectively an interpolation target when held out, whereas an anchor at a geometric extreme of the deployment loop is closer to an extrapolation target, a condition both a global similarity fit and a locally-propagating pose-graph correction handle less accurately. 
This effect is generally strengthened, not weakened, by pose-graph refinement: comparing each session/trajectory's coarse-only ratio to its refined ratio (Table~\ref{tab:collinearity}), the \{B,E\} advantage grows in three of the four pose-graph-corrected cases (e.g.\ UGV sparse, 0.75 coarse to 0.17 refined; UGV dense, 1.02 to 0.56), with UAV sparse the one exception (0.30 to 0.48, attributable to an already low error). 
This is consistent with the interpolation mechanism proposed above: because pose-graph refinement propagates anchor constraints locally along the trajectory rather than applying one rigid global correction, it should benefit most at points already well-bracketed by retained neighbours on both sides, precisely the condition that near-collinear midpoint anchors such as B and E satisfy by construction. 
The refinement step does not merely inherit the coarse fit's existing bias toward these anchors; it amplifies it.

Both aspects of this finding are relevant for deployment planning: geometric collinearity is a real, measurable property of an anchor placement, but it does not by itself predict poor individual-point accuracy under leave-one-out evaluation; anchor placement near the extremes of a deployment area is the more relevant risk factor.

\subsection{Cross-Platform Structural Agreement}
\begin{figure} [!ht]
\begin{center}
\begin{tabular}{c} 
\includegraphics[height=10cm]{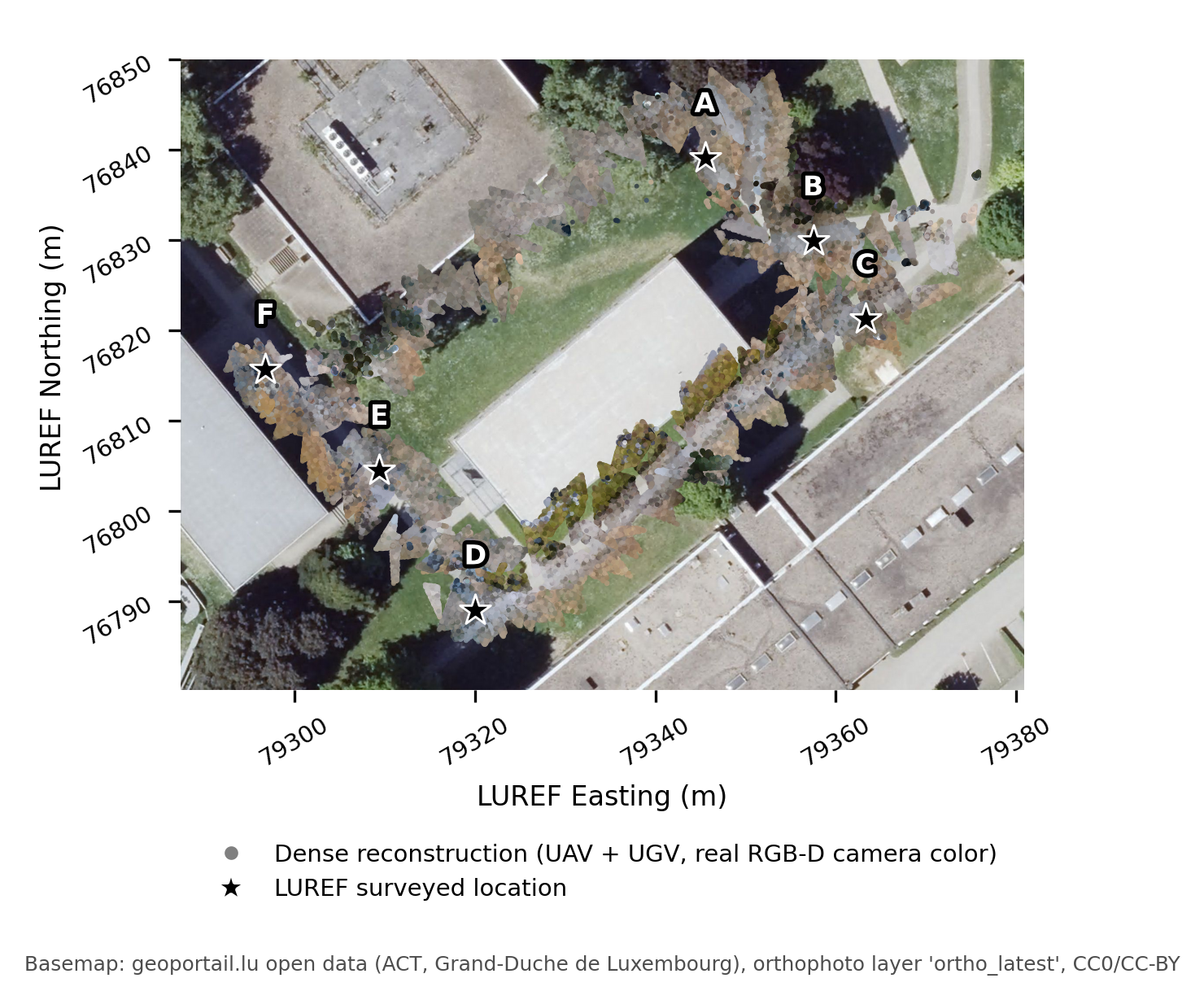}
\end{tabular}
\end{center}
\caption[fused map] 
{ \label{fig:fused_map} 
Dense reconstructions from the UAV and UGV sessions, each independently georeferenced to LUREF via its own marker anchors, fused into a single map without cross-session registration. Stars mark the six surveyed anchor locations.}
\end{figure}

Figure~\ref{fig:fused_map} shows the UAV and UGV sessions' dense reconstructions, each independently georeferenced via its own marker-anchor fit with no cross-session information used, fused into a single LUREF-referenced map. 
The two reconstructions fully overlap in spatial extent and visually trace the same building loop, passing through all six surveyed locations. 
Nearest-neighbor distance from the UAV reconstruction to the UGV reconstruction has a mean of 102.1\,cm and a median of 58.0\,cm, with 29.7\% of points agreeing within 30\,cm. 
This is moderate agreement, consistent with each platform's own georeferencing accuracy (Table~\ref{tab:correction_summary}) rather than contradicting it, and constitutes direct evidence that independently marker-anchored reconstructions from different platforms land in mutual spatial agreement without any explicit cross-platform registration step.

\subsection{Computational Cost}
Table~\ref{tab:cost} reports the computational cost of the two marker-dependent processing steps, measured on a single-threaded process (Intel Core i9-12900H, no GPU).

\begin{table}[h]
\centering
\caption{Computational cost of marker-dependent processing.}
\label{tab:cost}
\begin{tabular}{lc}
\toprule
Step & Cost \\
\midrule
Marker detection (per frame) & 12.4--12.6\,ms ($\sim$80\,fps) \\
Pose-graph correction (per session) & 18.6--233.7\,ms \\
\bottomrule
\end{tabular}
\end{table}

Detection cost is dominated by mask generation and candidate search, run on every frame regardless of marker presence, rather than by pose estimation itself. 
Pose-graph correction cost scales with trajectory sample count rather than anchor count and remains well under a quarter of a second even for the longest trajectory evaluated (1829 nodes). Neither step represents a computational barrier to practical deployment at this data scale.

\section{Discussion}
\label{sec:discussion}

\subsection{Marker Anchors as an Effective Correction Mechanism}
The central result of this work (Table~\ref{tab:correction_summary}) is that treating marker detections as active correction constraints, rather than as a single post-hoc reference frame, yields a substantial and consistent accuracy improvement across both trajectories.
Held-out generalization accuracy improves by 8--30\% over a coarse one-time alignment alone. 
Revisit self-consistency error is reduced by 97.9--99.1\% for three of the four session/trajectory combinations; the exception, UGV dense mapping, shows only a 15.5\% reduction because its raw, uncorrected self-consistency (12.9\,cm) was already close to the corrected result for the other cases, leaving comparatively little room for further improvement, in absolute terms it remains the most self-consistent of the four uncorrected trajectories by a wide margin. 
This validates the core premise of the proposed framework, that engineered, unambiguous anchor points can substitute for the shared structural overlap that heterogeneous platforms otherwise lack, and does so with a quantitative margin large enough to be practically, not just statistically, meaningful. 
Even a small number of anchors (six locations in this deployment) measurably constrains drift that would otherwise accumulate unbounded over a long-term session.

\subsection{Anchor Placement Matters More Than Anchor Count Alone}
Section~\ref{sec:results} presents two findings that together argue for treating anchor \emph{placement}, not only anchor \emph{count}, as a first-class deployment consideration.
First, accuracy degrades systematically as the effective anchor count is reduced (Table~\ref{tab:anchor_density}), for both trajectories, confirming the intuitive expectation that denser anchor coverage improves georeferencing quality. 
Second, anchors positioned near-collinearly with their neighbours (B, E) are consistently better predicted when held out than anchors at the geometric extremes of the deployment loop (Table~\ref{tab:collinearity}), despite being measurably closer to a degenerate configuration for constraining a rigid or similarity fit; this advantage is generally strengthened, not weakened, by pose-graph refinement, since refinement propagates anchor constraints locally and therefore benefits most at points already well-bracketed by retained neighbours on both sides. 
The reconciliation of these two findings is that leave-one-out accuracy at a given point depends on whether that point is interpolated between retained neighbours or extrapolated beyond them; near-collinear midpoint anchors are, by construction, favourable interpolation targets, and pose-graph correction reinforces rather than removes that advantage. 
For deployment planning, this suggests that the priority when placing a limited number of physical markers should be ensuring coverage at the spatial extremes of an area of interest, rather avoiding near-collinear placements, which this data shows to be comparatively less impacting regardless of whether coarse alignment or full pose-graph correction is used.

\subsection{Limits of a Single Calibrated Noise Model}
The pose-graph correction's odometry-edge noise model, calibrated once from directly observed drift on one trajectory, was found not to transfer reliably across sessions or trajectory types. 
For ICP odometry, a noise model calibrated from UAV's drift rate proved too tight for UGV's markedly faster drift, causing the optimizer to over-trust UGV's raw odometry edges. 
For dense mapping, the same pattern appeared in the opposite direction: a noise model calibrated from UAV's dense drift rate proved too loose for UGV's much slower dense drift, causing the optimizer to under-trust already-reliable odometry edges. 
In both cases the mismatch is not an implementation artifact but a genuine methodological limitation of calibrating this parameter once and reusing it: because the noise model governs how strongly the optimizer trusts anchor constraints relative to raw odometry, an uncalibrated mismatch systematically biases correction quality in a direction that is not obvious without independently measuring the trajectory's own drift characteristics first, and the direction of that bias depends on whether the true drift rate is faster or slower than the one used for calibration. 
We deliberately did not re-tune this parameter per result, since doing so would risk overfitting the reported accuracy figures to the evaluation itself rather than reflecting a procedure that would generalize to new
deployments. 
A practical implication is that any deployment of this correction method should include a short calibration phase, ideally using a revisit of a known point, as this study's protocol already provides, to establish a trajectory- and platform-specific noise model before relying on the correction's held-out accuracy.

\subsection{Cross-Platform Agreement}
The fused multi-platform reconstruction (Section~\ref{sec:results}, Figure~\ref{fig:fused_map}) demonstrates that two independently marker-anchored sessions, captured with the same handheld device under different simulated platform trajectories, land in consistent spatial agreement without any explicit cross-session registration step. 
The median 58\,cm nearest-neighbor agreement is meaningful evidence that the framework achieves cross-session georeferencing under GNSS-denied operation using pre-surveyed marker coordinates, but should be read alongside the underlying per-platform accuracy figures (Table~\ref{tab:correction_summary}) rather than as an independent, tighter bound: the fused map is built from each session's dense reconstruction, and its agreement is a direct consequence of, and is bounded by, each platform's own dense-trajectory georeferencing accuracy, not a separate validation of it.

\subsection{Practical Deployability}
The computational cost results (Section~\ref{sec:results}) indicate that neither marker detection nor anchor-based correction presents a barrier to practical use at the data scale evaluated here: detection runs comfortably above real-time frame rates on ordinary, single-threaded consumer hardware, and even the largest correction problem evaluated completes in well under a second.
Combined with the marker's imperceptibility and camouflage-matched design (Section~\ref{sec:imarker_design}), this supports the framework's motivating use case: a lightweight, concealable, computationally inexpensive georeferencing aid suitable for GNSS-denied defence and reconnaissance operations, rather than a laboratory-only proof of concept.

\subsection{Limitations}
Three limitations of the present evaluation should inform how its results are interpreted and what follow-on work would most usefully address. 
First, this study used a single physical marker relocated sequentially between surveyed locations, which precludes simultaneous multi-anchor detection and introduces placement uncertainty beyond the surveyed benchmark's own precision; a production deployment with multiple uniquely-encoded markers would remove both constraints and may reveal different accuracy characteristics, particularly under simultaneous rather than sequential anchor availability. 
Second, both evaluated trajectories were captured with a single handheld sensor rig under different motion profiles intended to emulate UAV and UGV behavior, rather than with physically distinct platforms; while this isolates the framework's core anchoring mechanism from platform-specific hardware variation, validation on genuine UAV and UGV hardware remains necessary before the cross-platform claim can be considered fully established. 
Third, marker detection coverage across a full session remains dependent on the marker being within the sensor's field of view for a sufficient time at each location; deployment protocols should account for this dependency explicitly rather than assuming incidental visibility during transit is sufficient.

\section{Conclusion and Future Works}
\label{sec:conclusion}
This paper presented a framework using imperceptible, camouflage-matched CSR-based fiducial markers as active correction anchors for georeferencing platform-emulating trajectories and dense reconstructions during GNSS-denied operation using pre-surveyed marker anchors. 
Using six surveyed locations and a single relocated marker, anchor-based pose-graph correction reduced trajectory self-consistency error by 97.9--99.1\% for three of the four evaluated session/trajectory combinations, and improved held-out georeferencing accuracy by 8--30\% over one-time alignment alone, with within-cluster marker-pose repeatability of 15.3--18.8\,cm and negligible computational cost. Independently anchored dense reconstructions from two platform-emulating sessions achieved consistent spatial agreement without explicit cross-session registration, and anchor placement was shown to matter more through spatial coverage than through avoiding near-collinear configurations, an effect that pose-graph refinement was
found to strengthen rather than weaken.

Future work should validate the framework with multiple simultaneously-deployed markers on physically distinct UAV and UGV hardware, and develop trajectory-specific noise-model calibration to remove the manual tuning step identified in Section~\ref{sec:discussion}.

\acknowledgments % equivalent to \section*{ACKNOWLEDGMENTS}       
This work was partially funded by the Fonds National de la Recherche of Luxembourg (FNR) under the project DEFENCE22/IS/17800397/INVISIMARK.

% References
\bibliography{report} % bibliography data in report.bib
\bibliographystyle{spiebib} % makes bibtex use spiebib.bst

\end{document}